\documentclass[11pt]{article}

\usepackage[final]{acl}

\usepackage{times}
\usepackage{latexsym}
\usepackage{amsmath}
\usepackage{amsfonts}
\usepackage{enumitem}
\usepackage{graphicx}
\usepackage{xcolor}
\usepackage{booktabs}
\usepackage[table]{xcolor}
\usepackage{tikz}
\usepackage{multirow}

\usepackage[T1]{fontenc}

\usepackage[utf8]{inputenc}

\usepackage{microtype}

\usepackage{inconsolata}

\title{Interpretable Difficulty-Aware Knowledge Tracing in \\ Tutor-Student Dialogues}

\author{
Shuyan Huang \quad Alexander Scarlatos \quad Jaewook Lee \quad Andrew Lan \\
University of Massachusetts Amherst \\
\texttt{\{shuang, ajscarlatos, jaewooklee, andrewlan\}@umass.edu}
}

\begin{document}
\maketitle

\begin{abstract}

Recent advances in large language models (LLMs) have led to the development of AI-powered tutoring systems that provide interactive support via dialogue. To enable these tutoring systems to provide personalized support, it is essential to assess student performance at each turn, motivating knowledge tracing (KT) in dialogue settings. However, existing dialogue-based KT approaches often ignore question difficulty and rely on opaque LLM latent representations, hindering accurate and interpretable prediction. In this work, we propose an interpretable difficulty-aware conversational KT framework that leverages LLMs to explicitly model student knowledge state and the difficulty of tutor-posed tasks at each dialogue turn. The framework incorporates the original question and the next tutor-posed task to estimate the student's knowledge state and the difficulty of the upcoming turn. It further integrates Item Response Theory to map LLM outputs into student ability and question difficulty parameters, enabling interpretable prediction of student performance grounded in cognitive theories of learning. We evaluate the framework on two tutor-student dialogue datasets. Quantitative and qualitative results show that our framework outperforms existing KT baselines, meanwhile generating interpretable outputs consistent with cognitive theory \footnote{Our code and data are available at \url{https://github.com/umass-ml4ed/Difficulty-Aware-DialogKT}}.

\end{abstract}

\section{Introduction}

Tutoring, which provides tailored instruction and feedback to students in an interactive manner, has been shown to effectively improve learning outcomes \cite{nickow2020impressive, nye2014autotutor}. With the advanced natural language understanding and generation capabilities of large language models (LLMs), researchers have developed an increasing number of LLM-based tutoring systems, such as Khanmigo \cite{khan2023khanmigo} and LiveHint \cite{carnegielearning2024livehint}. 
Recent work has further explored how to improve LLM-based tutors by training them to optimize student learning outcomes and by steering them toward diverse tutor personas that reflect different scaffolding, feedback, and affective-support styles \cite{scarlatos2025training, lee2026letting}.
These systems can significantly reduce the workload of human tutors and provide scalable support for a large number of students.

\begin{figure}
    \centering
    \includegraphics[width=1\linewidth]{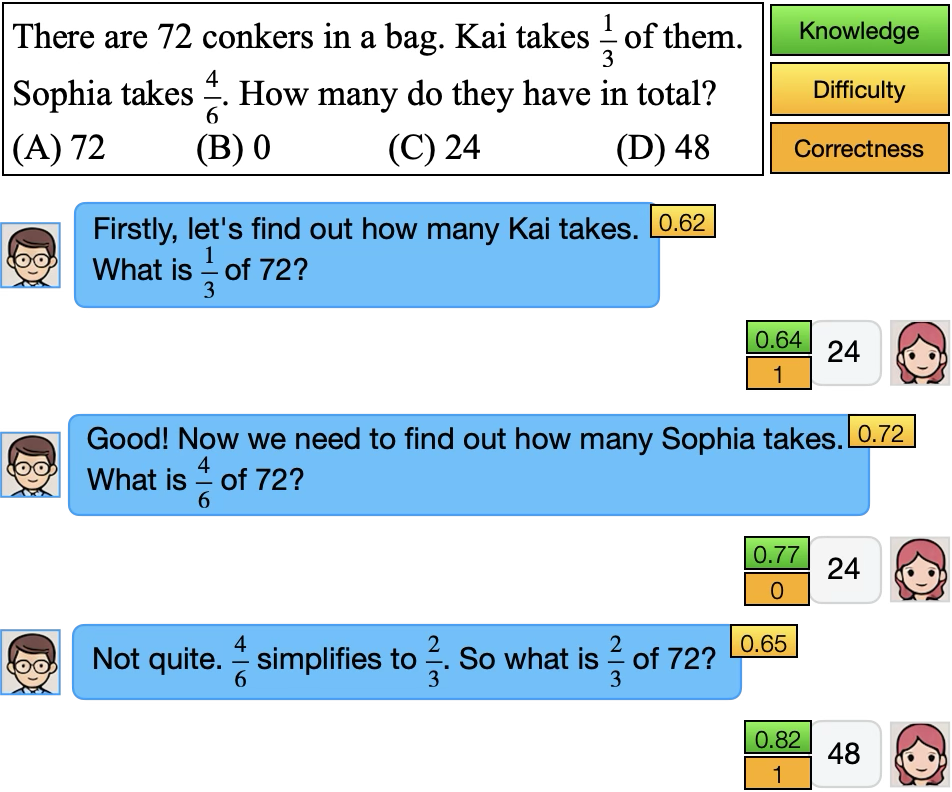}
    \caption{An example dialogue illustrating a student's learning trajectory as the tutor adapts difficulty while scaffolding the math problem.}
    \label{fig:illustration}
\end{figure}

To provide personalized pedagogical support, it is crucial to assess students' mastery of concepts based on their responses throughout the dialogue. The process of modeling students' knowledge states over time to predict their future performance is known as knowledge tracing (KT). Recently, with the development of AI-powered tutoring systems, KT has been extended to dialogue-based settings to predict whether students can correctly answer subsequent tutor-posed tasks \cite{scarlatos2025exploring, wang2025training}.

However, existing KT approaches in dialogue settings have two important limitations. First, prior studies do not explicitly model question difficulty. In practice, the correctness of student responses is not solely determined by their knowledge state but is also substantially influenced by question difficulty \cite{shen2022assessing}. Even students with high knowledge mastery may fail to answer very difficult questions. For example, as shown in Figure \ref{fig:illustration}, although the student has a relatively high estimated knowledge state (e.g., 0.64), the student still answers the next tutor-posed task incorrectly when the question difficulty increases (e.g., from 0.62 to 0.72). Prior studies that model student responses solely based on students' knowledge states may yield inaccurate predictions \cite{scarlatos2025exploring}. Second, the reasoning process underlying model predictions of students’ future performance remains opaque. Most current approaches rely on LLMs with high-dimensional latent representations that lack interpretability. Consequently, it is difficult to derive psychologically meaningful explanations that align with cognitive theories of learning. This lack of transparency may undermine tutors’ trust in model predictions, hindering the adoption of LLMs as effective tutoring tools. As illustrated in Figure \ref{fig:illustration}, even when the student answers a more difficult task incorrectly, the estimated knowledge state continues to increase, reflecting gradual learning dynamics consistent with cognitive theories \cite{newell1981mechanisms, anderson2000learning}.

\subsection{Contributions}

To address the aforementioned challenges, we propose an interpretable difficulty-aware conversational KT framework built upon an LLM-based architecture that explicitly models both student ability and tutor-turn difficulty. Specifically, we prompt an LLM with the question and dialogue context to estimate student knowledge, and further prompt the model with the tutor turn to estimate the difficulty of the tutor-posed task. Furthermore, to bridge the gap between model predictions and cognitive theories of learning, we integrate principles from Item Response Theory (IRT). More specifically, we feed the estimated ability and difficulty into an IRT model to predict student correctness, enabling fine-tuning of the LLM using observed correctness labels. This design aligns model predictions with cognitive theory and improves interpretability by providing transparent reasoning based on the relationship between student ability and task difficulty. To evaluate the effectiveness of our approach, we conduct comprehensive experiments on two tutor-student dialogue datasets. Both quantitative and qualitative results demonstrate the superiority of our framework in student ability and difficulty estimation throughout tutoring dialogues.

\section{Problem Statement}

Practically, a dialogue often starts when a student requests assistance from a tutor when they cannot successfully solve a math question/problem. Formally, we represent each dialogue $\mathcal{D}$ as an ordered sequence of alternating tutor turns $t$ and student turns $s$, i.e., $\mathcal{D} = (s_0, t_1, s_1,\ldots, t_N, s_N)$, where $N$ denotes the total number of turn pairs in the dialogue and $s_0$ is included when the student initiates the dialogue. In cases where the same speaker produces several consecutive utterances, we merge them into a single tutor turn or student turn to maintain the alternating turn structure. Following \citet{scarlatos2025exploring}, each dialogue can be treated as a sequence of formative assessments, where each tutor turn $t_j$ (for turn index $j$) acts as an assessment question, and the corresponding student turn $s_j$ represents a correct or incorrect response. This perspective allows us to use tutor-student dialogues to model student learning over time. Each tutor turn with pedagogical intent (i.e., designed to assess student knowledge) is associated with one or more related knowledge components (KCs), $\mathcal{C}_j=\{c_j^k\}_{k=1}^{K_j}$, where $K_j$ is the number of KCs associated with turn $t_j$. For student turns that follow these tutor turns, we define the correctness of a student turn $s_j$ as $r_j \in \{0, 1\}$. In this way, we formulate the dialogue-based KT task as predicting the correctness $\hat{r}_{j+1}$ of the student’s next response with the given question $q$, historical dialogue $(t_1,s_1,...,t_j,s_j)$, and next tutor-posed task $t_{j+1}$, which can be formally defined as:
\begin{equation}
   \hat{r}_{j+1} \sim f_\Theta(q,t_1,s_1, ...,t_j,s_j, t_{j+1}),
   \label{eq:problem_formalizaiton}
\end{equation}
where $\boldsymbol{\Theta}$ denotes the learnable parameters of the framework, which are optimized to predict students' future performance.

\section{Approach}

\begin{figure*}
    \centering
    \includegraphics[width=1.0\textwidth]{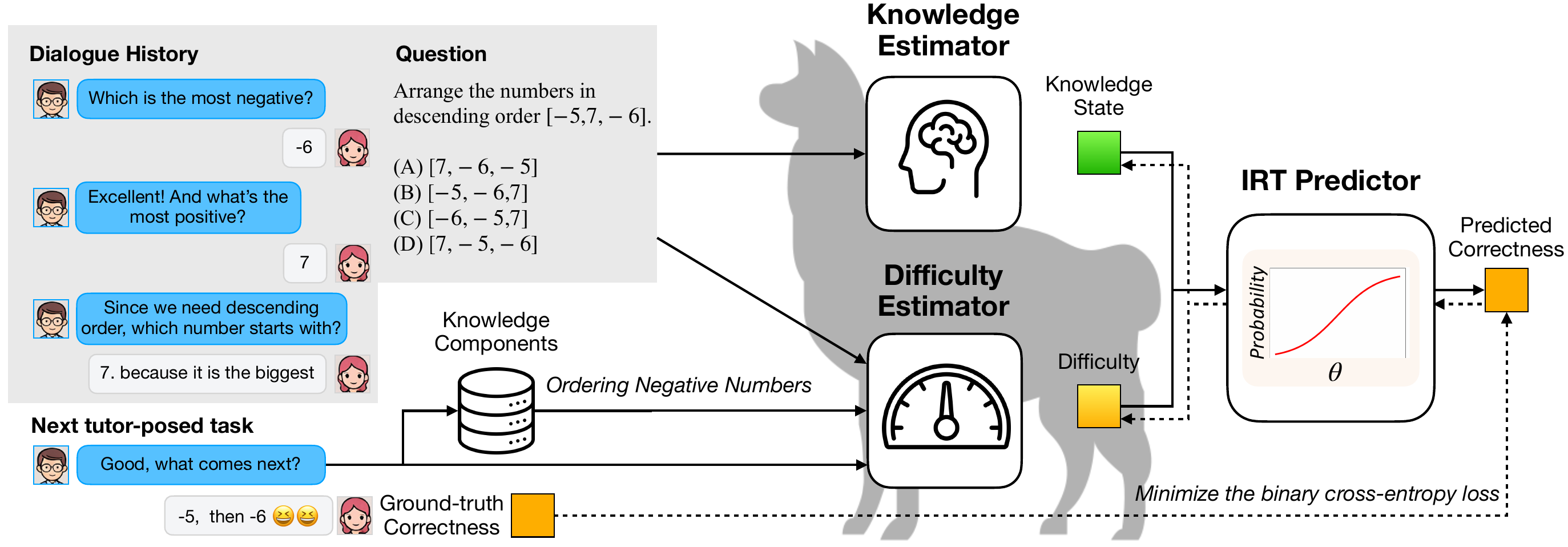}
    \caption{Overview of our framework with three modules: a knowledge estimator, a difficulty estimator, and an IRT-based predictor that jointly model student knowledge and task difficulty to predict correctness on the next task.}
    \label{fig:overview}
\end{figure*}

In this section, we detail our framework that consists of three modules: (1) a \emph{knowledge estimator} that extracts latent knowledge representations from both the textual content of the question and the tutor-student dialogue history; (2) a \emph{difficulty estimator} that computes the difficulty of the tutor-posed task based on the question content, the dialogue history and content of the current turn; and (3) an \emph{IRT-based predictor} that integrates the estimated knowledge state and difficulty representation to provide interpretable predictions of whether the student will correctly respond to the tutor-posed task. We illustrate our full framework in Figure~\ref{fig:overview}, and show our prompts in Appendix~\ref{sec:appendix-prompts}.

\subsection{Knowledge Estimator}

We now introduce the knowledge estimator module of our framework, which estimates student knowledge states for each turn in a dialogue. At each turn, the student's subsequent performance depends on both the original question being solved and the dialogue history up to the current turn. The original question provides necessary context, while the dialogue history reveals the student's evolving knowledge state, including observed errors or misconceptions. To effectively represent the textual information, we leverage the natural language understanding capabilities of LLMs to jointly encode the question text and dialogue history.
To adapt general-purpose LLMs to the educational domain, we further fine-tune the model on dialogue-based tutoring data.
We construct a prompt $I_{j+1}^\text{K}$ representing turn $j+1$, where we instruct the LLM to generate a one-word prediction for the student's current ability level. Prior LLM-based KT work categorizes students' knowledge states into discrete levels, e.g., good/fail \citep{li2025explainable}. Inspired by this classification design, we use ``GOOD'' and ``BAD'' as positive and negative verbalizer tokens following prompt-based classification methods \citep{schick2021exploiting}. Rather than sampling text from the LLM, we construct a continuous ability estimate by comparing their next-token logits. We first extract the LLM's next-token output logits corresponding to these two verbalizer tokens:

\[
z_{j+1}^{\text{GOOD}},\; z_{j+1}^{\text{BAD}} = \text{LLM}(I_{j+1}^\text{K}).
\]

We convert these logits into a scalar ability estimate by taking their difference, defining the student's ability at the next turn as:
\[
\theta_{j+1} = z_{j+1}^{\text{GOOD}} - z_{j+1}^{\text{BAD}},
\]
where $\theta_{j+1}$ represents the student's latent knowledge state at turn $j+1$.

\subsection{Difficulty Estimator}

We now introduce the difficulty estimator module of our framework, which estimates the difficulty of each tutor-posed task in a dialogue. Difficulty estimation is recognized as a fundamental component in classical assessment settings, where accurately modeling item difficulty improves the reliability of student evaluation \cite{scarlatos2025smart}. In dialogue-based tutoring, each interaction is typically centered around an original question that the student is attempting to solve with tutor assistance. This original question is associated with an inherent difficulty determined by its textual content and cognitive complexity. As the dialogue progresses, the tutor can actively pose tasks with different difficulty in subsequent turns. Hence, to more accurately predict student performance at each turn, we estimate the difficulty of each tutor-posed task by jointly considering the original question text, dialogue context, and the next tutor-posed task. Similar to the knowledge estimator, we construct a prompt $I_{j+1}^\text{D}$ representing turn $j+1$, where we instruct the model to generate a one-word prediction for the difficulty level of the tutor-posed task. Again, we extract the LLM's output logits, now corresponding to the ``HARD'' and ``EASY'' verbalizer tokens:

\[
z_{j+1}^{\text{HARD}},\; z_{j+1}^{\text{EASY}} = \text{LLM}(I_{j+1}^\text{D}).
\]
We convert the logits into a scalar difficulty estimate, defining the difficulty of the tutor-posed task:

\[
d_{j+1} = z_{j+1}^{\text{HARD}} - z_{j+1}^{\text{EASY}},
\]
where $d_{j+1}$ represents the latent difficulty of the tutor-posed task at turn $j+1$.

\subsection{IRT-based Predictor}

In prior LLM-based dialogue KT approaches, it is very challenging to explain the LLMs' parameters and their decision-making processes when predicting student response correctness. Therefore, to align LLMs' predictions with models in psychometric measurement, we introduce an IRT-based prediction layer that jointly predicts student response correctness while providing interpretable estimates of student knowledge and tutor turn difficulty. IRT, particularly the one-parameter logistic (1PL) model, also known as the Rasch model \cite{rasch1993probabilistic}, is widely used in educational assessment to model the probability of a correct response based on student ability and task difficulty. In the 1PL IRT model, student ability and task difficulty are represented as latent variables, and the probability of correctness increases as the positive difference between ability and difficulty grows. This formulation provides interpretability since both student knowledge and task difficulty are explicitly represented as meaningful cognitive quantities. Given the estimated ability $\theta_{j+1}$ and task difficulty $d_{j+1}$, the probability that a student will answer correctly at turn $j+1$ is defined as:
\begin{equation*}
\hat{r}_{j+1} = \frac{1}{1+\exp(-\alpha(\theta_{j+1}-d_{j+1}))},
\end{equation*}
where $\alpha$ is a learnable scalar that controls the sensitivity of the probability to differences between ability and difficulty. This parameter plays a role similar to the discrimination parameter in the two-parameter logistic (2PL) IRT model \cite{birnbaum1968some}, enabling the model to adjust how strongly performance depends on the gap between ability and difficulty. Please note that we additionally experimented with modeling $\alpha$ as a learnable vector to capture KC-specific discrimination, analogous to the 2PL model. 
However, this setting led to degraded performance in our experiments, likely due to overparameterization. Therefore, we adopt a scalar $\alpha$ in our final framework.

\subsubsection{Model Optimization}

To improve predictive accuracy, we fine-tune the LLM on correctness in observed tutor-student dialogues. We optimize the learnable parameters of the model by minimizing the binary cross-entropy loss between the ground-truth correctness of student responses $r_{j+1}$ and the predicted probabilities $\hat{r}_{j+1}$ produced by the IRT-based prediction layer:
\begin{equation*}
\mathcal{L} \!= \!- \! \sum(r_{j+1}\text{log}\hat{r}_{j+1}+(1-r_{j+1})\text{log}(1-\hat{r}_{j+1})).
\end{equation*}

\section{Experiments}
\subsection{Datasets}

To evaluate the effectiveness of our framework, we conduct experiments on two datasets as follows:

\paragraph{Question-Anchored Tutoring Dialogues (QATD$_\text{2k}$)} QATD$_\text{2k}$ is the largest open-source real-world tutoring dialogue dataset~\cite{zent2025piivot}. The dataset is collected from one-on-one conversations between tutors and students on the Eedi mathematics learning platform. Each dialogue begins when a student requests assistance in solving a math multiple-choice problem. After data preprocessing, the dataset contains 1,573 and 393 dialogues in the training set and test set, respectively. We use the LLM-generated KC and response correctness annotations collected in prior work~\cite{scarlatos2026simulatedstudentstutoringdialogues}.

\paragraph{MathDial} MathDial is a tutoring dataset grounded in
multi-step math reasoning problems between GPT-3.5-simulated students and crowd workers acting as tutors~\cite{macina2025mathtutorbench}. Each dialogue begins with an incorrect student solution and the tutor guides the student to identify and correct their errors. The dataset contains 2,235 and 588 dialogues in the training set and test set, respectively. We use the LLM-generated KC and response correctness annotations collected in prior work~\cite{scarlatos2025exploring}.

\subsection{Baselines}

We compare our framework against two types of representative KT baselines: LLM-based models and deep learning-based models.

\subsubsection{LLM-based models}
These models leverage LLMs to model student learning behavior by encoding question content and dialogue history to predict student responses.
\begin{itemize}[leftmargin=*]
\item \textbf{LLMKT}: leverages textual question and conversation information in tutor--student dialogues using an LLM-based architecture, which is a previous state-of-the-art dialogue-based KT model \cite{scarlatos2025exploring}. 
\end{itemize}

\subsubsection{Deep learning-based models}
These models employ conventional deep neural networks to capture student knowledge states from sequences of their historical interactions, which consist of question tags, KC tags and binary responses, rather than LLMs.


\begin{itemize}[leftmargin=*]
\item \textbf{DKT}: leverages an LSTM to sequentially model students' learning processes and predict their future performance \cite{piech2015deep}.
\item \textbf{DKVMN}: leverages a static key memory matrix to encode relationships among KCs and a dynamic value memory matrix to predict students' knowledge mastery levels \cite{zhang2017dynamic}.
\item \textbf{SAINT}: leverages a Transformer-based architecture to model students' sequential learning interactions over time \citep{choi2020towards}.
\item \textbf{AKT}: leverages a monotonic attention mechanism to explicitly capture the temporal distance between current questions and students' historical interactions \citep{ghosh2020context}.

\item \textbf{simpleKT}: leverages a Rasch-inspired architecture to model question-specific variations and dot-product attention for modeling students' sequential learning \citep{liu2023simplekt}.

\end{itemize}

\subsection{Metrics and Experimental Setup}

To quantitatively evaluate KT models in dialogue settings, we report Area Under the Curve (AUC) and Accuracy. We compute these metrics over all correctness labels except for the first label in each dialogue, since no prior observations are available for KT prediction at the initial step.



For our framework, we adopt Llama-3.1-8B-Instruct \citep{grattafiori2024llama} as the base model. We implement the model using the HuggingFace Transformers library \citep{wolf2020transformers} and fine-tune it with LoRA \citep{hu2022lora} on NVIDIA RTX L40 GPUs for efficient adaptation.


We perform a grid search to select hyperparameters for each model, choosing the checkpoint with the highest validation AUC to evaluate the test set. For LLM-based models, we search learning rates $\in \{1\times10^{-5}, 5\times10^{-5}, 1\times10^{-4}, 2\times10^{-4}\}$ and LoRA ranks $\in \{16, 32\}$ and fixing the LoRA scaling factor $\alpha=16$. The effective batch size is 64 via gradient accumulation, with gradient clipping at 1.0. We train the models for 5 epochs using BFloat16 mixed precision. For other KT baselines, we search learning rates $\in \{1\times10^{-4}, 2\times10^{-4}, 5\times10^{-4}, 1\times10^{-3}, 2\times10^{-3}, 5\times10^{-3}\}$ and embedding sizes $ \in \{8, 16, 32, 64, 128, 256, 512\}$. The batch size is 64 and the total epoch is 100, the remaining hyperparameters follow the default settings in pyKT \cite{liu2022pykt}. We train all models with AdamW optimizer with a weight decay of $1 \times 10^{-2}$.

\section{Experimental Results}

\begin{table}[tpbh]
\centering
\setlength{\tabcolsep}{3pt}
\begin{tabular}{lcc|cc}
\toprule
Model 
& \multicolumn{2}{c}{QATD$_\text{2k}$}
& \multicolumn{2}{c}{MathDial} \\

\cmidrule(lr){2-3}
\cmidrule(lr){4-5}

& Accuracy 
& AUC 
& Accuracy 
& AUC \\

\midrule

\rowcolor{gray!20}
\multicolumn{5}{c}{Deep Learning-based Models} \\
\midrule

DKT       
& $60.19$ 
& $54.73$ 
& 59.75 
& 63.14 \\

DKVMN     
& $58.24$ 
& $51.79$ 
& 56.42 
& 60.64 \\

SAINT     
& $60.71$ 
& $53.76$ 
& 59.80 
& 63.22 \\

AKT       
& $60.88$ 
& $54.39$ 
& 59.75 
& 63.37 \\

simpleKT  
& $61.62$ 
& $56.24$ 
& 58.94 
& 61.90 \\

\midrule

\rowcolor{gray!20}
\multicolumn{5}{c}{LLM-based Models} \\
\midrule

LLMKT 
& $\underline{64.24}$ 
& $\underline{64.89}$ 
& $\underline{68.64}$ 
& $\underline{75.99}$ \\

Ours 
& $\mathbf{64.29}$ 
& $\mathbf{65.25}$ 
& $\mathbf{68.82}$ 
& $\mathbf{76.59}$ \\

\bottomrule

\end{tabular}

\caption{Student dialogue turn correctness prediction performance on QATD$_\text{2k}$ and MathDial datasets. Best results are shown in \textbf{bold} and second-best results are \underline{underlined}.}

\label{tab:overall}

\end{table}

\subsection{Quantitative Results}

Table \ref{tab:overall} shows the overall performance in terms of AUC and Accuracy for all models. We have the following observations: First, our proposed framework achieves the best overall performance and improves upon the previous state-of-the-art LLMKT model by explicitly modeling student ability, item difficulty, and combining them with IRT-based prediction. Specifically, our framework improves AUC from 64.89 to 65.25 (+0.36) on QATD$_\text{2k}$ and from 75.99 to 76.59 (+0.60) on MathDial. These improvements suggest that explicitly modeling question difficulty provides additional useful information for monitoring student learning progress. Although the performance gains are relatively modest, our framework introduces interpretable outputs, which we further analyze model interpretability in Section \ref{sec:learning_curve}; Second, in general, LLM-based methods consistently outperform deep learning-based KT models in terms of both AUC and Accuracy, demonstrating the advantage of leveraging pretrained LLMs in dialogue-based KT settings. We attribute this improvement to the ability of pretrained LLMs to model rich textual content in dialogues, enabling us to better track student learning after fine-tuning on response correctness labels. Third, among deep learning-based KT methods, DKT achieves comparable performance with those complex architectures such as SAINT and AKT in both datasets. These results suggest that increasing architectural complexity does not necessarily yield significant performance gains. Instead, simpler architectures may generalize more effectively in dialogue-based KT settings, particularly under limited-data conditions (fewer than 2,500 dialogues), where complex models are more susceptible to overfitting.



\subsection{Qualitative Analysis: Difficulty Analysis}

\begin{figure}[tp]
    \centering
    \includegraphics[width=1\linewidth]{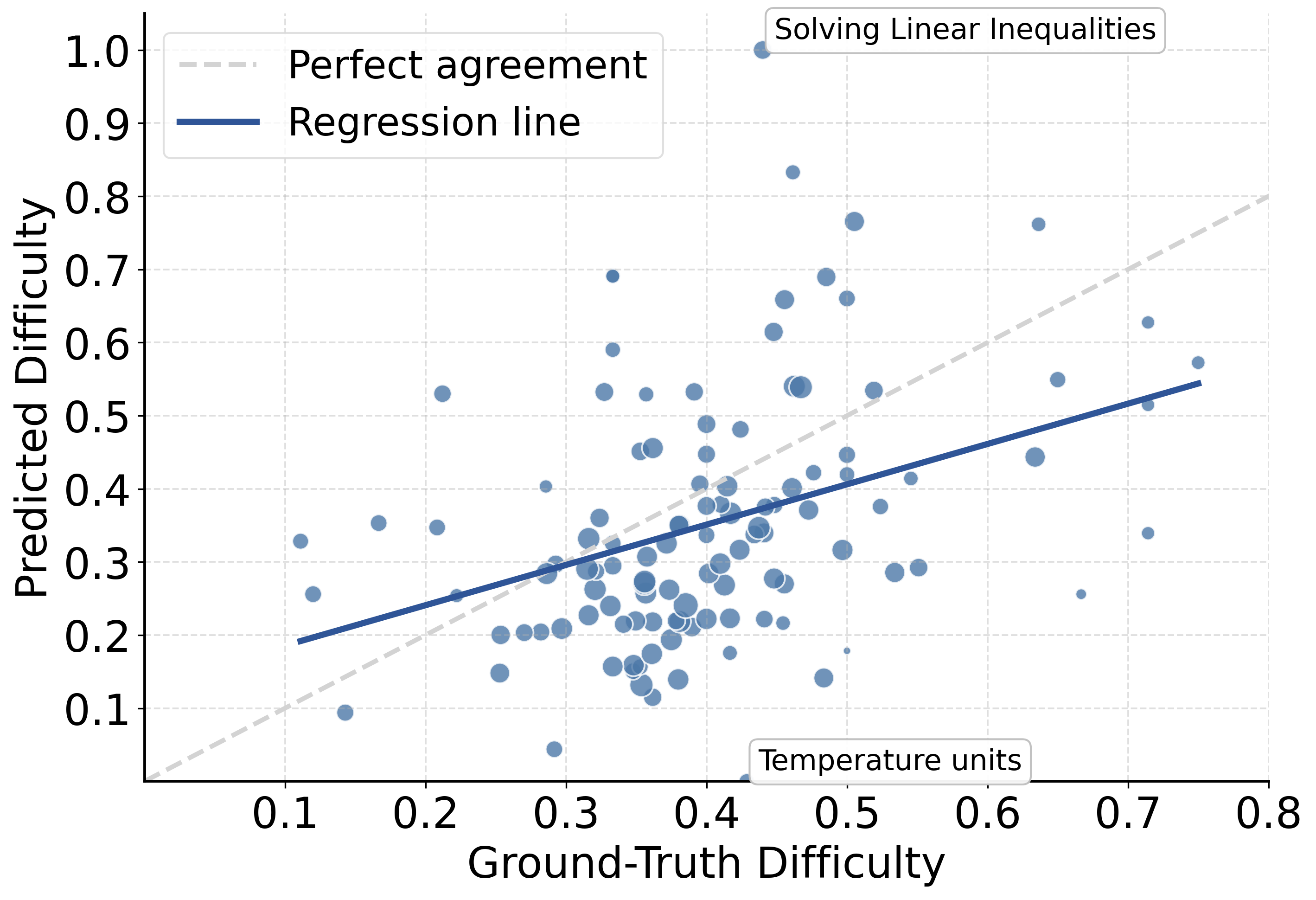}
    \caption{Comparison of predicted and ground-truth KC difficulty across KCs in the test set of the QATD$_\text{2k}$ dataset. Each point represents one KC. Point size is proportional to the number of samples associated with each KC. The dashed gray line indicates perfect agreement, while the solid blue line shows the fitted linear regression trend.}
    \label{fig:difficulty}
\end{figure}

To show that the learned difficulty parameters are indeed meaningful and interpretable, we measure the consistency between predicted KC difficulty and ``ground-truth'' KC difficulty. We use QATD$_\text{2k}$ instead of MathDial for this analysis, since QATD$_\text{2k}$ is collected from real-world educational environments and therefore better captures authentic tutor-student behaviors. Similar to \cite{lee2024difficulty}, we follow classical test theory to define the ground-truth KC difficulty based on empirical correctness statistics as $ 1 - \frac{N_{\text{correct}}}{N_{\text{total}}}$, i.e., the portion of incorrect responses on this KC \cite{brown2013classical}. To ensure a fair comparison, we normalize the predicted difficulty values using min--max normalization. As illustrated in Figure~\ref{fig:difficulty}, the fitted regression line follows a similar increasing trend to the perfect agreement line, indicating that the predicted difficulty generally increases as the ground-truth difficulty increases. To quantitatively support this observed trend, we compute the Pearson correlation between predicted and ground-truth difficulty values across all KCs \citep{benesty2009pearson}. Results show a moderate positive correlation ($r = 0.368$, $p < 0.001$), suggesting that the learned difficulty parameters capture meaningful empirical difficulty patterns in real-world tutoring data across KCs.



\begin{figure*}
    \centering
    \includegraphics[width=1.0\textwidth]{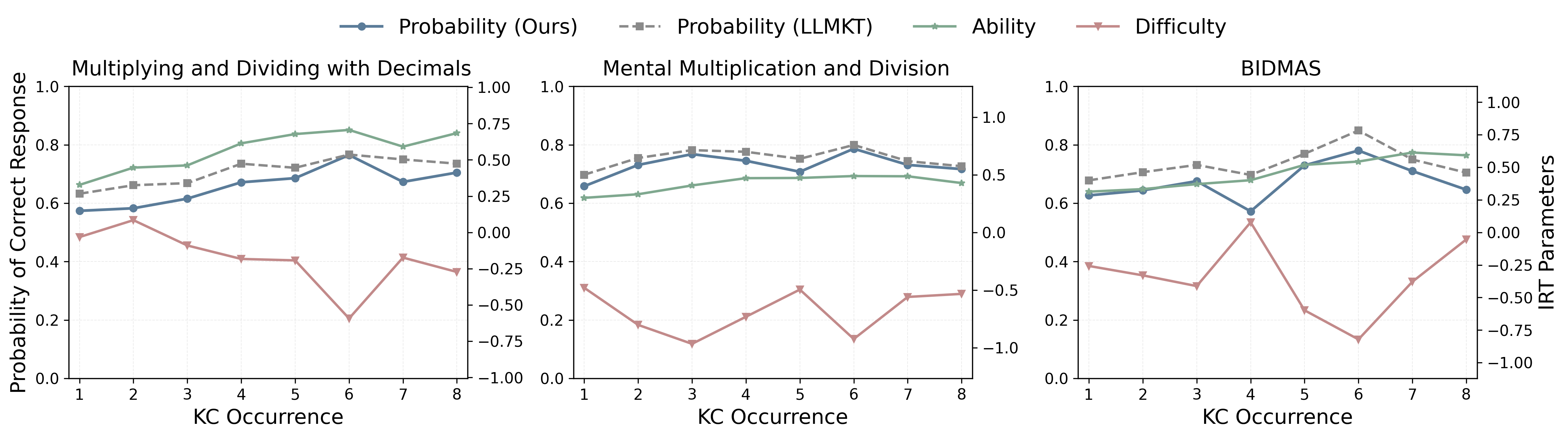}
    \caption{Learning trajectories for the three most frequent KCs in the QATD$_{\text{2k}}$ dataset.}
    \label{fig:learning_curve}
\end{figure*}

\subsection{``Learning Curve'' Analysis}
\label{sec:learning_curve}

A common approach to evaluate KT quality is to examine whether predicted learning dynamics align with cognitive theories such as the power law of practice \cite{snoddy1926learning}. Following prior work \cite{shi2023kc, scarlatos2025exploring}, we visualize learning trajectories of each KC. We use QATD$_\text{2k}$ instead of MathDial for this analysis, since QATD$_\text{2k}$ is collected from real-world educational environments and therefore better captures authentic tutor-student behaviors. Specifically, for each KC, we group student interactions by the number of prior opportunities to practice that KC, where an ``occurrence'' refers to each time the KC appears in a tutor-posed task within a dialogue. For each occurrence index, we compute the average predicted mastery across all corresponding dialogue turns in the test set. The resulting curves therefore illustrate how the estimated mastery evolves as students encounter the same KC multiple times. 

We examine plots of the top three most frequent KCs, as they contain sufficient samples and provide the most intuitive visualization of learning trends. As shown in Figure~\ref{fig:learning_curve}, LLMKT, which directly treats predicted correctness probability as knowledge mastery, exhibits noticeable oscillations across KC occurrences. The predicted probability decreases even after repeated practice, suggesting unstable representations of student mastery. In contrast, our framework separates ability estimation from correctness prediction, resulting in smoother and more stable learning trajectories. The estimated student ability consistently increases with repeated KC occurrences across all three KCs, reflecting an increase in knowledge mastery that aligns with the cognitive theories of gradual learning \cite{ritter2002learning}. 



Furthermore, fluctuations in predicted correctness probability from our framework are largely associated with changes in tutor-posed task difficulty. For example, temporary drops in predicted correctness probability often coincide with increases in estimated task difficulty, while the underlying ability continues to increase. 
Such fluctuations are expected in dialogue-based tutoring settings, where task difficulty may decrease when the tutor provides hints or scaffolding, and increase when the tutor switches to a different, more difficult part of the question or presses the student for a complete solution. Hence, our framework reflects realistic interactions between student learning progress and task difficulty. Similar trends observed across multiple KCs further demonstrate that our framework produces more stable and cognitively plausible learning dynamics than probability-based mastery estimation models such as LLMKT. We also provide learning trend visualizations at the dialogue level in Appendix~\ref{sec:appendix_learning}. 



\subsection{Qualitative Analysis: Case Study}

\begin{table*}[ht!]
    \centering
    \small
\usetikzlibrary{decorations.pathreplacing}

\begin{tabular}{p{4.5cm} p{5.5cm} llll}
\toprule
$q$ & $t_j$ & $d_j$ & $s_j$ & \textbf{$\theta_j$} &  $r_j$\\
\midrule
\multirow{6}{=}{%
\begin{minipage}[t]{\linewidth}
\raggedright
Which question could the bar model represent?\\[4pt]

\begin{center}
\begin{tikzpicture}[x=1.1cm,y=1.1cm]
    \draw (0,0) rectangle (1,0.8);
    \draw (1,0) rectangle (2,0.8);
    \draw (2,0) rectangle (3,0.8);
    \draw (3,0) rectangle (4,0.8);

    \node at (0.5,0.4) {?};
    \node at (1.5,0.4) {?};
    \node at (2.5,0.4) {?};
    \node at (3.5,0.4) {?};

    \draw[decorate,decoration={brace,mirror,amplitude=5pt}]
        (0,-0.08) -- (3,-0.08)
        node[midway,yshift=-12pt] {15};
\end{tikzpicture}
\end{center}

A: Calculate $\frac{3}{4}$ of $15$\\
B: $\frac{3}{4}$ of $\square = 15$\\
C: Calculate $\frac{1}{3}$ of $15$\\
D: $\frac{4}{3}$ of $\square = 15$
\end{minipage}
}
& what is the bar model split up into? & -- & 15 & \textcolor{red!50!black}{0.224} & 0 \\\cmidrule(l){2-6}
& and in terms of the whole bar model, how much does the 15 cover? & 0.671 & half. & \textcolor{red!50!black}{0.475} & 0 \\\cmidrule(l){2-6}
& not quite! How many boxes is the bar split into? & 0.360 & 4 & \textcolor{green!50!black}{0.531} & 1 \\\cmidrule(l){2-6}
& okay good, and how many does the 15 span across? & 0.447 & All of the bar & \textcolor{green!50!black}{0.605} & 0 \\\cmidrule(l){2-6}
& as in the bottom arrow. How many boxes is it under? & 0.440 & 1 & \textcolor{green!50!black}{0.628} & 0 \\\cmidrule(l){2-6}
& not quite, can you see that the 15 and the arrow go across 3 boxes? & 0.405 & -- & -- & -- \\
\bottomrule
\end{tabular}
    \caption{Case study of an adaptive tutoring dialogue illustrating how task difficulty ($d_j$) and student ability ($\theta_j$) evolve over successive interactions in the QATD$_\text{2k}$ dataset. The tutor dynamically scaffolds the problem by decomposing it into simpler sub-tasks following incorrect responses, while progressively maintaining or increasing difficulty after correct answers, leading to a gradual improvement in the student’s knowledge state.}
    \label{tab:case_study}
\end{table*}

We further conduct a qualitative case study to examine how the difficulty of tutor-posed tasks and the student's knowledge state evolve in a dialogue in the QATD$_\text{2k}$ dataset. Specifically, we analyze one representative dialogue from the test set. As shown in Table~\ref{tab:case_study}, the tutor dynamically adjusts the difficulty of the posed tasks based on the student's responses. When the student fails to answer correctly, the tutor decomposes the original question into simpler sub-tasks, leading to a decrease in the estimated task difficulty, from 0.671 to 0.360. In contrast, when the student answers correctly, the tutor proceeds with follow-up questions that maintain or slightly increase the difficulty level, moving the dialogue forward. Meanwhile, the student's knowledge state demonstrates a generally increasing trend across the dialogue. Notably, even when incorrect responses occur, the estimated student ability continues to improve gradually with the tutor's guidance, from 0.224 to 0.628 across the full dialogue. This observation highlights that learning progress may still occur despite temporary errors, as scaffolding and intermediate feedback from the tutor help consolidate the student's KC understanding. Overall, this case study illustrates that our framework captures realistic tutoring dynamics, where task difficulty adapts to student performance while student ability evolves steadily through guided interaction. 



\section{Related Work}

\subsection{Student Modeling in Dialogues}

Recently, with the emergence of LLM-based tutoring systems, several studies have explored KT in dialogue settings. \citet{scarlatos2025exploring} leveraged the textual content in dialogues to estimate student knowledge. \citet{wang2025training} proposed a workflow that integrates KT with turn-by-turn verification to improve coding tutoring \cite{wang2025training}. However, these approaches overlook the modeling of question difficulty, which plays a critical role in KT.  On the other hand, our approach introduces a difficulty-aware dialogue-based KT framework that explicitly aligns the student knowledge state and tutor-posed task difficulty, enabling fine-grained modeling of student state evolution. Beyond KT, recent studies have increasingly explored LLM-based student simulation in educational dialogue settings. Prior work has demonstrated that LLMs can be used to simulate realistic student learning behaviors, including the development of learning curves and misconception patterns \cite{schmucker2024ruffle,jin2024teach}. Other studies simulate student behaviors conditioned on demographic, behavioral, and personality traits \cite{markel2023gpteach, li2025type}, as well as varying learning goals and trajectories \cite{sharma2024designing}.


\subsection{Traditional Knowledge Tracing}

Traditional KT models capture a student's learning behavior in a sequence of her/his historical interactions, which consist of question tags, KC tags and binary responses \cite{chen2023improving, liu2023enhancing, huang2023towards, huang2026dual, liu2025prompt, guo2026improving}. For example, \citet{piech2015deep} introduced the first DKT model, utilizing an LSTM to estimate knowledge mastery. \citet{zhang2017dynamic} proposed a static key memory matrix to store KC relationships and a dynamic value memory matrix for predicting knowledge mastery levels. To incorporate the difficulty factor to enhance the knowledge state estimation, \citet{ghosh2020context} introduced a scalar difficulty parameter to control how far a question deviates from the underlying concept during question representation learning. \citet{shen2022assessing} established the relationship between student knowledge state and question difficulty level to improve KT performance. \citet{liu2024question} replaced question tags with difficulty levels to facilitate the translation of model predictions into interpretable skill-level knowledge states, further highlighting the importance of difficulty modeling in KT.

\section{Conclusion and Future Work}

In this work, we introduce an interpretable difficulty-aware conversational KT framework that jointly models student ability and question difficulty in tutor-student dialogues. Our framework aims to capture fine-grained difficulty signals and provide more interpretable predictions of student performance, connecting LLM-based modeling with cognitively motivated assumptions. Quantitative results show modest improvements over existing KT baselines, suggesting the potential benefit of explicitly modeling both student ability and task difficulty. Empirical qualitative analyses provide further evidence for the interpretability of the learned parameters and indicate the value of incorporating an IRT predictor. In particular, the learned difficulty parameters show positive correlations with empirical difficulty statistics, suggesting that the model captures meaningful and cognitively plausible difficulty patterns across KCs. Additionally, the estimated ability trajectories often exhibit smooth growth patterns broadly consistent with gradual learning, while fluctuations in predicted correctness tend to correspond to changes in tutor-posed task difficulty.

There are many avenues for future work. First, since we explicitly consider the difficulty factor in dialogue-based KT settings, we will further explore more effective methods for capturing fine-grained difficulty signals from tutoring dialogues and investigate how task difficulty evolves dynamically during tutoring interactions. Second, for student knowledge state estimation, we plan to leverage advanced reasoning capabilities and reinforcement learning techniques to improve the accuracy and robustness of the estimation. In particular, future work will explore how multi-step reasoning can ensure consistency across dialogue turns and how reinforcement learning methods can refine estimation strategies based on long-term prediction performance. Third, we plan to explore misconception-aware conversational KT by detecting and tracking students’ misconceptions from dialogue interactions. By integrating misconception signals with estimated knowledge states and difficulty parameters, the framework may provide more fine-grained and interpretable student performance prediction.

\section*{Acknowledgments} This work is partially supported by Renaissance Philanthropy under the learning engineering virtual institute (LEVI) initiative and the NSF under grant 2237676. 

\section*{Limitations}

There are several practical limitations to our work. First, since our work is the first to explicitly estimate the difficulty of tutor turns in dialogues, there are no existing difficulty prediction baselines we can compare against. Second, we only conduct experiments on math dialogues, and do not investigate other domains such as language learning or computer science. However, this is largely because KT in dialogues is currently only well-established in math. Finally, while our framework and other LLM-based approaches are significantly more accurate than traditional KT approaches, they are much more computationally expensive too. An important line of future work will be improving the efficiency of LLM-based approaches while maintaining high predictive accuracy.

\section*{Ethical Considerations}

There are both potential societal benefits and risks associated with our work. Accurately modeling student ability and difficulty in dialogues is a necessary step to facilitating automated assessments in the dialogue setting. This can potentially benefit students through more accurate assessment and subsequently more targeted learning actions. However, as is the case with many other advances in educational AI, further automating assessments creates a risk of replacing teachers' jobs. Our intention is for educational AI tools such as ours to be used in conjunction with teachers to improve outcomes for students, rather than be used to replace teachers. Finally, there is a risk of bias in our framework, as is common in all AI methods, due to inherent biases in training data. For instance, it is possible that students from populations that are less represented in training data are given less accurate assessments, therefore risking worse outcomes for those students. It is important that educational AI tools are thoroughly screened for bias before deployment with real students.




\bibliography{acl}

\appendix

\section{Prompts}
\label{sec:appendix-prompts}

\subsection{Prompt for Knowledge Estimation}

\begin{quote}
\textit{You are an experienced math teacher. Given a dialogue where the student is working through a multiple-choice diagnostic math question, based on the student's responses and demonstrated understanding in the dialogue, classify the student's current ability level. Respond with exactly one token from: GOOD BAD.}
\begin{verbatim}
[BEGIN QUESTION]
Question Text
[END QUESTION]

[BEGIN DIALOGUE]
Dialogue Context
[END DIALOGUE]
\end{verbatim}
\end{quote}

Specifically, the input prompt $I^\text{K}$ consists of the following three key components:
\begin{itemize}[leftmargin=*]
\item \textbf{Task Definition}: A natural language instruction that defines the task and specifies the expected output format for the LLM.
\item \textbf{Question Text}: The diagnostic question presented to the student, typically enclosed within structured markers \texttt{[BEGIN QUESTION]} and \texttt{[END QUESTION]}.
\item \textbf{Dialogue Context}: The historical teacher--student interaction associated with the question, enclosed within markers \texttt{[BEGIN DIALOGUE]} and \texttt{[END DIALOGUE]}.
\end{itemize}

\subsection{Prompt for Difficulty Estimation}

\begin{quote}
\textit{You are an experienced math teacher. Given a dialogue where the student is working through a multiple-choice diagnostic math question, based on the question content and required knowledge components, classify the question's difficulty level. Respond with exactly one token from: HARD EASY.}
\begin{verbatim}
[BEGIN QUESTION]
Question Text
[END QUESTION]

[BEGIN DIALOGUE]
Dialogue Context
[END DIALOGUE]
\end{verbatim}

\begin{verbatim}
[BEGIN CURRENT TEACHER TURN]
Teacher-posed Task
[END CURRENT TEACHER TURN]
\end{verbatim}

\begin{verbatim}
[BEGIN KC]
KC Names
[END KC]
\end{verbatim}

\end{quote}

\begin{figure}
    \centering
    \includegraphics[width=0.8\linewidth]{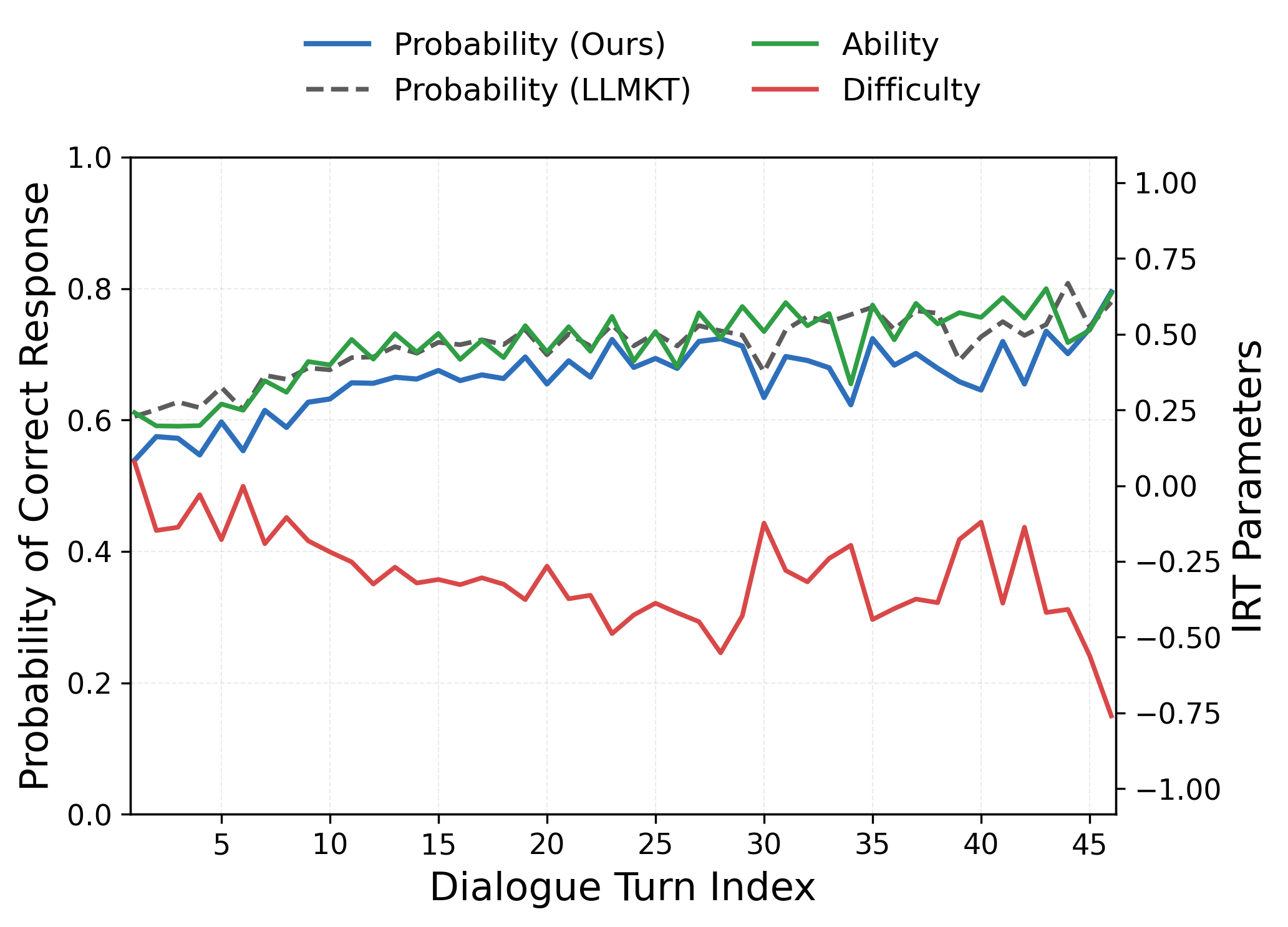}
    \caption{Learning trajectories for dialogues in the QATD$_{\text{2k}}$ dataset.}
    \label{fig:learning_curve_dialog}
\end{figure}

Specifically, the input prompt $I^\text{D}$ consists of the following two key components:

\begin{itemize}[leftmargin=*]
\item \textbf{Teacher-posed Task}: The next teacher-posed question that the student is expected to respond to is enclosed within markers [BEGIN CURRENT TEACHER TURN] and [END CURRENT TEACHER TURN], representing the immediate task used for predicting the student's knowledge state.

\item \textbf{KC Names}: The associated KCs with the next tutor-posed task enclosed within markers [BEGIN KC] and [END KC]. The model must infer from prior interactions whether the student has mastered the required knowledge.
\end{itemize}

\section{Additional ``Learning Curve'' Analysis}

We provide learning curve analysis based on dialogue level in Figure \ref{fig:learning_curve_dialog}.

\label{sec:appendix_learning}

\end{document}